\documentclass{article} 
\usepackage{iclr2027_conference,times}
\usepackage{wrapfig}

\usepackage{amsmath,amsfonts,bm}

\def\eqref#1{equation~\ref{#1}}

\def\1{\bm{1}}

\DeclareMathAlphabet{\mathsfit}{\encodingdefault}{\sfdefault}{m}{sl}
\SetMathAlphabet{\mathsfit}{bold}{\encodingdefault}{\sfdefault}{bx}{n}

\usepackage{hyperref}
\usepackage{url}
\usepackage{graphicx}
\usepackage{enumitem}
\usepackage{booktabs}

\title{Artemis: Geometry-Grounded Multi-Agent Driving World Models with Shared 3D State and Progressive Memory Update}

\author{
Sitian Shen\textsuperscript{\rm 2}$^{*}$,
Jiuming Liu\textsuperscript{\rm 1}$^{*\ddagger}$,
Mengmeng Liu\textsuperscript{\rm 3}\thanks{These authors contributed equally, their order was determined by drawing lots. }\ \ , 
Yian Wang\textsuperscript{\rm 4},
 Michael Ying Yang\textsuperscript{\rm 5},\\
\bf Francesco Nex\textsuperscript{\rm 3},
Hao Cheng\textsuperscript{\rm 3}, 
Daniele De Martini \textsuperscript{\rm 2},
Ayush Tewari\textsuperscript{\rm 1}\thanks{Corresponding Author. $\ddagger$ Project Lead.},
Per Ola Kristensson\textsuperscript{\rm 1} \\
\textsuperscript{1} University of Cambridge\quad
\textsuperscript{2} University of Oxford \quad
\textsuperscript{3}University of Twente\quad \\
\textsuperscript{4} National University of Singapore\quad
\textsuperscript{5}University of Bath \\
\texttt{sitian.shen@eng.ox.ac.uk liujiuming123@gmail.com m.liu-1@utwente.nl} \\
}

\iclrfinalcopy 
\begin{document}

\maketitle
\lhead{Preprint}

\begin{figure}[h]
\centering
\includegraphics[width=\linewidth]{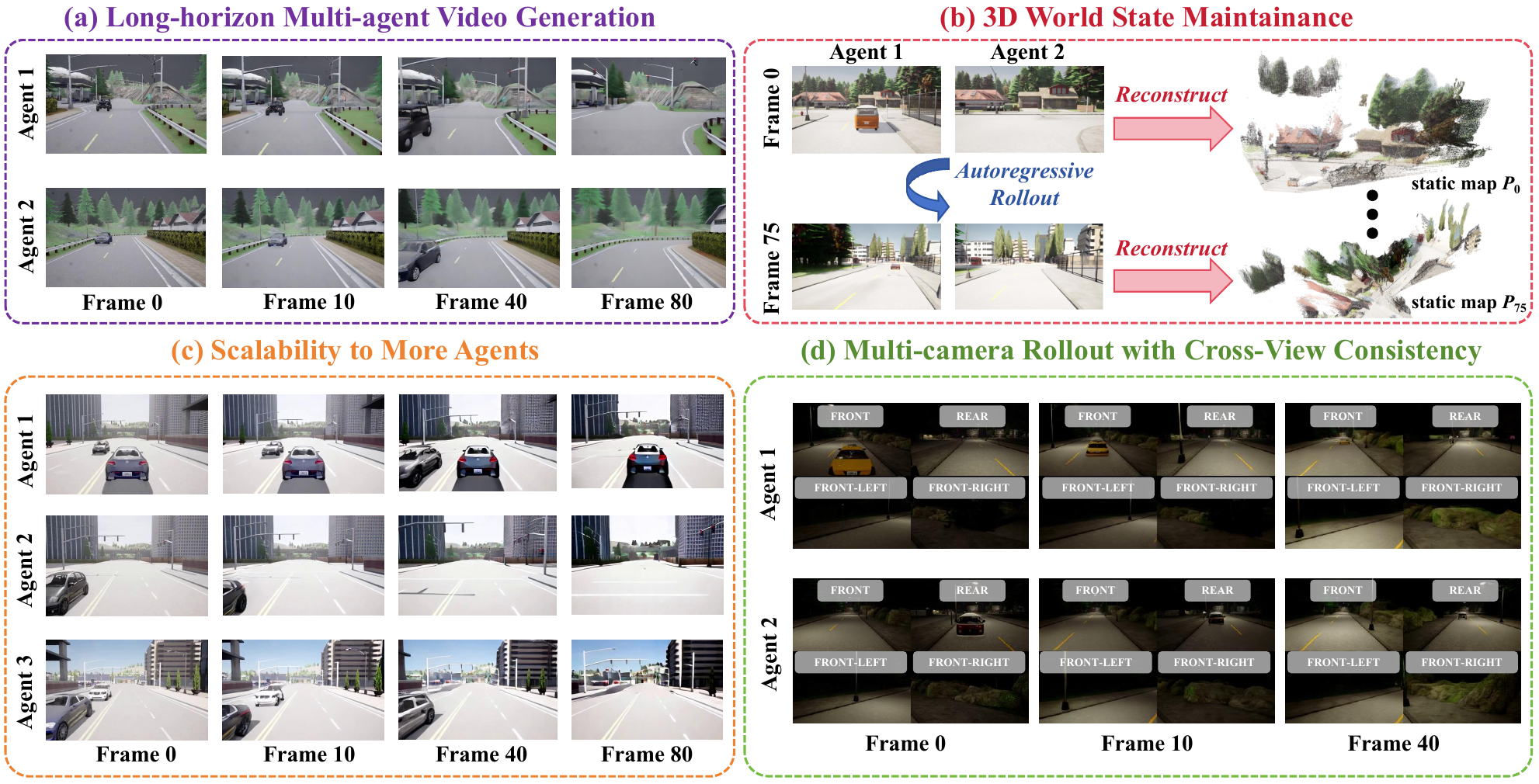}
\vspace{-6mm}
\caption{\textbf{Artemis: a geometry-grounded multi-agent world model.} (a) Our method can generate long-horizon future videos given initial observations and following actions. (b) Distinct from existing multi-agent methods, we explicitly construct and dynamically update a 3D world state using static point map. (c) With our curated MA-CARLA dataset consisting of various interaction modes, Artemis supports scalable agent numbers beyond two. (d) Furthermore, due to our designed explicit 3D constraints, multi-camera videos can also be created with high cross-view consistency.}
\label{fig:head}
\end{figure}

\begin{abstract}
Recent video world models have witnessed the paradigm shift from single-agent to multi-agent involvements, which can reveal more complicated dynamics and cross-agent interaction in the real world. However, existing approaches commonly adopt implicit inter-agent communications via cross attention, which lack explicit geometry constraints and unified 3D state, thereby leading to poor multi-view consistency and struggling with recovering out-of-sight agents. In addition, most of them assume a static background, failing to represent uncontrolled background dynamics. To address these problems, we propose Artemis: a geometry-grounded multi-agent world model with explicit memory sharing. An explicit 3D world map is reconstructed from multi-agent observations to enforce a unified 3D state across agents, offering high cross-view consistency. Specifically, an action-guided geometric injection module is developed to simultaneously render decomposed foreground-background control maps, which are then injected into a diffusion transformer through a designed GeoAdapter block. Compared to previous methods assuming static-only background, our GeoAdapter can also distinguish uncontrolled non-agent dynamics, which are conditioned on their own multi-frame history positions to provide consistent motion cues. Keyframes selected from progressive video rollouts are used to progressively update the reconstructed 3D world maps. To effectively capture complex dynamic patterns, we curate a novel dataset sampled from the CARLA simulator called MA-CARLA, with scalable agent numbers, physical plausibility, and abundant interaction modes. Extensive experiments demonstrate the superiority of our proposed method in terms of visual fidelity and cross-view consistency in the generated videos. In addition, our Artemis can support simultaneous multi-modal rollouts with both 2D video and 3D point map maintenance, scale to scenarios beyond two agents, and flexibly switch between single-camera or multi-camera setting. Project page: \url{https://liujiuming123.github.io/Artemis/}.
\end{abstract}

\section{Introduction}
Video world models \citep{ha2018world,po2025long,wu2025video,liu2025worldweaver,zhao2025drivedreamer} can predict future environmental dynamics given historical observations and actions from users or agents, which have recently achieved remarkable success in autonomous driving \citep{wang2024drivedreamer,zhao2025drivedreamer,liu2026driveva,liu2026universe}, embodied AI \citep{lu2025gwm,seo2023multi,huang2026pointworld}, and immersive game engines \citep{yu2025gamefactory,zhang2025matrix}. However, early methods are mostly constrained by a single-agent perspective, which cannot reflect the intrinsic multi-agent interactions and complex dynamics in the real world.

To effectively extend to multi-agent interactive world models, some recent approaches attempt to achieve inter-agent information exchange in a shared world. As a pioneering work, Solaris \citep{savva2026solaris} curates a novel multiplayer dataset in gaming scenes and introduces shared self-attention layers for cross-agent correlation. Subsequent methods further improve it with scalability in agent numbers \citep{liu2026gamma,wu2026multiworld}, accurate agent positional encoding \citep{sun2026prisma,liu2026gamma}, and world state alignment \citep{wu2026multiworld,hu2026metaworld}. 

However, these multi-agent approaches commonly struggle with three challenges: (1) \textbf{Lack of explicit geometric constraint for cross-view consistency}: The physical environment and action-triggered events should evolve consistently and be faithfully reflected from all agents' perspectives. This requires a globally consistent understanding of geometric structures and visual appearances across all agents. MultiWorld \citep{wu2026multiworld} proposes an action attention and reweighting module to maintain a global state. However, their inter-agent association is still established on implicit cross attention, lacking sufficient geometric cues to ensure multi-view consistency. (2) \textbf{Limited memory sharing}: Basically, diverse agents execute their actions and exist in the same shared world, which means that historical observed frames across agents may share high overlap, especially in scenarios when another agent revisits the same location. Therefore, sufficient historical memory sharing is crucial to further strengthen a unified world state and facilitate cross-agent information reuse. However, existing methods \citep{savva2026solaris,wu2026multiworld,hu2026metaworld} mostly fail to consider effective cross-agent geometry sharing beyond the current timestamp. (3) \textbf{Over-simplified scene complexity and dynamics}: Previous methods typically evaluate the multi-agent interaction in the synthetic gameplay scenario where dynamics are quite simplified to reveal the true world evolution, leading to uncertain generalization capabilities in scaling agent complexity, memorizing out-of-sight dynamics, and handling severe occlusions in complex world distributions. 

To address these bottlenecks, we propose Artemis, a geometry-grounded multi-agent world model with shared explicit 3D state and memory. Rather than implicit-only agent association, our method constructs a 3D global point cloud-based world map with cross-view consistency to address the first challenge. Explicit geometric constraints are then rendered as control maps and then injected into Diffusion Transformer (DiT) through GeoAdapter, enabling global-aware and view-invariant video generation for each agent. Specifically, we decompose the scene into static background, action-controlled foreground agents, and uncontrolled background dynamics. Static geometry and controlled agents' proxies provide background and foreground RGB and depth control maps, respectively, while uncontrolled non-agent dynamics further require multi-frame historical control maps providing consistent motion tendency for future synthesis.

 For the second bottleneck, we further introduce two novel designs to facilitate effective cross-agent memory sharing: (1) \emph{World map sharing.} We progressively select keyframes revealing significantly unobserved areas from new generated videos across all agents to update a global world map. In this case, different agents can explicit share a consistent 3D world state with multi-view matching priors. (2) \emph{Relative pose sharing.} During autoregressive rollout, persistent agent proxies are propagated using cross-agent relative poses and updated from rolling-out videos. Another benefit of maintaining relative pose is that, when an unrepresented controlled agent is expected to enter the view, a pose-dependent visibility test can initialize its proxy. To address the third challenge, we curate a novel dataset from the CARLA simulator in complex urban scenarios called MA-CARLA, which provides scalable multi-camera sampling, abundant interaction modes, physical plausibility, and diverse environmental dynamics. 


Overall, the key contributions of our paper are as follows:
\begin{itemize}
    \item We propose a novel multi-agent world model called Artemis with strong explicit geometric constraints and cross-agent memory sharing. Compared to prior methods that only leverage implicit feature exchange, our method recovers and dynamically maintains synchronized 3D world states, offering superior cross-view consistency. 
    \item Hybrid geometry-pose memory sharing provides complementary memories for disentangled statics and dynamics. An explicit world map memory is designed by progressively reconstructing static background from new generated videos which are integrated across agents. Additionally, relative-pose visibility checks can support proxy initialization of suddenly emerging vehicles.
    \item A novel dataset is designed based on the CARLA simulator, which enables physical plausibility, diverse inter-agent interaction modes, multi-camera sampling, and complicated environmental dynamics.
    \item Extensive experiments demonstrate that our Artemis surpasses prior methods in terms of visual fidelity, cross-view consistency, and uncontrolled background dynamics. Notably, our Artemis can support both single-camera and multi-camera setting, and also simultaneously roll out multi-modal driving scenes with consistent 2D and 3D representations.
\end{itemize}

\section{Related Work}
\textbf{Interactive Video World Models.} World models \citep{ha2018world} can predict future environmental dynamics given historical observations and actions. With rapid advances in recent video diffusion transformers, there is increasing research interest resorting to action-conditioned video generation backbones \citep{wan2025wan} for world modeling. Recently, interactivity has become a crucial property in world models where users can immersively navigate, manipulate, and personalize their created worlds in a multi-round paradigm. As a milestone, Genie \citep{bruce2024genie} compresses the video features into latent space with VAE and then learn actions unsupervisedly, supporting minute-level video generation and interactive exploration. Subsequent world models focus on achieving long-horizon consistency \citep{po2025long,huang2026live,wang2026matrix}, effective memory design \citep{wu2025video,xiao2025worldmem,hong2025relic,chen2026out}, or real-time rollout \citep{chen2026context,huang2025self,zhu2026sana}. However, most of video world models enforce a single-agent formulation with a first-person or third-person perspective, with limited agent numbers and exploration flexibility \citep{liu2026towards}.  

\textbf{Multi-Agent Video World Models.} To capture inter-instance interactions, MultiVerse \citep{enigma2025multiverse} proposes the first multiplayer gaming world model by trivially concatenating each agent's video and action tokens along the channel dimension for denoising together. Solaris \citep{savva2026solaris} further designs a shared self-attention and per-player ID embeddings. Nevertheless, its per-slot embeddings break the permutation-symmetry requirement for each agent. Moreover, Solaris suffers from poor generalization to arbitrary agent numbers due to quadratic-complexity full attention. To address these problems, Gamma-World \citep{liu2026gamma} introduces a regular simplex in rotary angle space to achieve equidistant agent embeddings and sparse hub attention to decrease computational costs. MultiWorld \citep{wu2026multiworld} constructs the testbed on the ItTakesTwo game and develops adaptive weighting in multi-agent action space. To strengthen view consistency, it also introduces VGGT \citep{wang2025vggt} as reconstruction priors to represent the 3D-aware global state. MetaWorld \citep{hu2026metaworld} observes that the collection of multi-view data is prohibitively expensive, thereby decomposing the camera ego-motion and object trajectories for data scaling. MultiGen \citep{po2026multigen} breaks down the function in diffusion game engines into memory module, observation module, and dynamic module. Similar to our insight, Prisma-World \citep{sun2026prisma} also views scene geometry as significant cues in 3D-aware consistent video generation. However, it only injects the relative camera pose into DiT for geometrical awareness, without explicit 3D maps.

\begin{figure*}[t]
\centering
\includegraphics[width=1.0\linewidth]{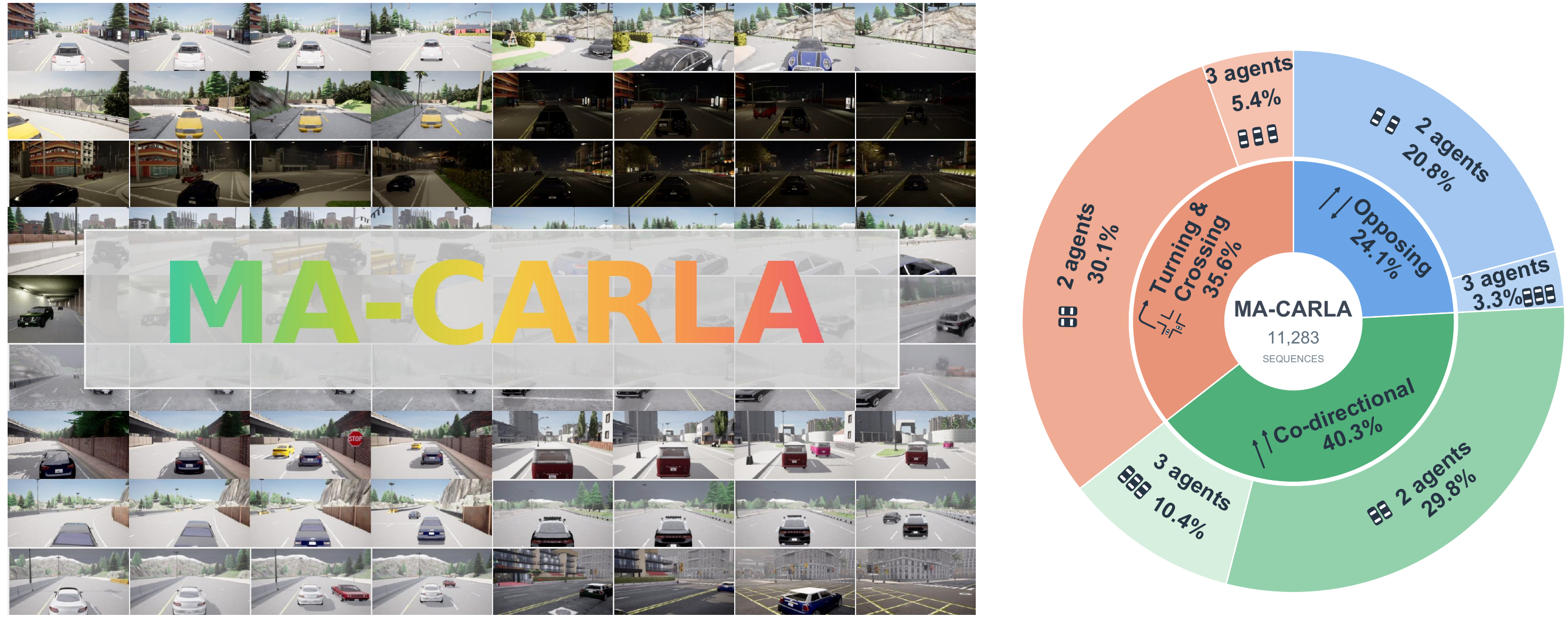}
\vspace{-6mm}
\caption{\textbf{MA-CARLA dataset.} We visualize various curated samples in our MA-CARLA dataset, and also calculate the distributions of diverse interaction modes and agent numbers.}
\vspace{-3mm}
\label{fig:dataset}
\end{figure*}

\section{MA-CARLA Dataset}
\label{sec:ma_carla}


\subsection{Data Collection}

We construct \textbf{MA-CARLA}, a synchronized multi-agent driving dataset rendered in CARLA~\cite{dosovitskiy2017carla}. It contains 11,283 rendered sequences derived from 2,478 unique synchronized trajectories across eight towns and 17 weather presets. The dataset includes 6,020 two-agent sequences with static backgrounds, 3,382 two-agent sequences with dynamic backgrounds containing non-agent actors, and 1,881 three-agent sequences with static backgrounds.

\begin{wrapfigure}{r}{0.55\textwidth}
    \vspace{-12pt}
    \centering
    \includegraphics[width=0.54\textwidth]{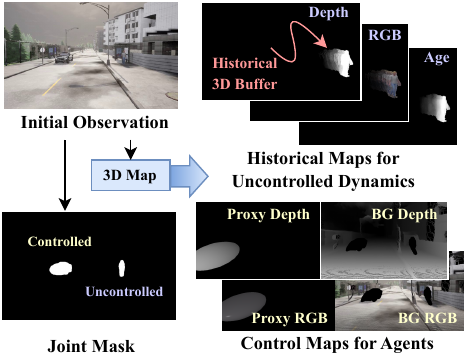}
    \vspace{-10pt}
    \caption{{Construction of control maps.}}
    \vspace{-12pt}
    \label{fig:control_map}
\end{wrapfigure}
Candidate trajectories are generated using rule-based templates for opposing, following, and overtaking configurations, with route variants that may include turns. After trajectory-level filtering, accepted motions are replayed under different weather conditions while preserving agent motion and synchronization. Each rendered sequence contains 256 synchronized timestamps. Each vehicle is configured with four selected views---front, front-left, front-right, and rear---at $800\times600$ resolution, together with motion annotations and camera calibration. Across the two-agent sequences, MA-CARLA provides approximately 19.3M available RGB images. Fig.~\ref{fig:dataset} visualizes representative samples and summarizes post-hoc motion groups extracted from recorded trajectories. A sequence may belong to multiple motion groups, yielding 17,790 group assignments in total. 
Fig.~\ref{fig:dataset} reports their distribution by motion group and agent count. Appendix~\ref{app:ma_carla} provides further details on data collection, grouping criteria, camera configurations, and data splits.

\subsection{Data Preprocessing}

We preprocess synchronized CARLA videos into geometry-aware conditioning signals. For each agent--camera streaming sequence, LingBot-Map~\citep{chen2026geometric} is used to estimates dense depth, intrinsics, and camera poses in a consistent local coordinate system. The same estimates are used for point lifting, proxy fitting, and control rendering to preserve the non-metric scale. 
Grounded-SAM2~\citep{ravi2025sam,liu2024grounding} extracts candidate dynamic masks, associated with either action-controlled agents or uncontrolled actors using CARLA metadata and calibrated projections. We exclude both categories from static-map fusion. Controlled agents are fitted as full-covariance 3D Gaussian proxies, while uncontrolled actors are stored as time-stamped observations rather than assigned future trajectories as in Fig.~\ref{fig:control_map}. Artemis then renders a unified 24-channel control tensor consisting of background RGB/depth, proxy RGB/depth, a synthesis mask, uncontrolled-history RGB/depth, and an age mask within a certain historical window.

\section{Artemis}

\subsection{Preliminaries}
\label{inference}
\textbf{Problem Formulation.} We investigate multi-agent interactive world model in an autoregressive geometry-anchored video generation paradigm which can model joint future probabilistic distributions given sequences of history observations and actions for each agent as:
\begin{equation}
    \mathcal{D}_{\theta}\left(\{\mathbf{F}^{n}_{t}\}_{n=1}^{N}, \mathbf{P}_{t}|\{\mathbf{O}^{n}_{< t},\mathbf{A}_{\leq t}^{n}\}_{n=1}^{N}\right),
\end{equation}
where $\mathbf{O}^{n}_{t}$ indicates the history observation and $\mathbf{A}^{n}_{t}$ denotes the input action at timestamp $t$ for the agent $n$. Past observations (memory) together with input actions will jointly formulate the conditional guidance for the generator $\mathcal{D}_{\theta}$, generating future frames $\mathbf{F}^{1}_{t}, ..., \mathbf{F}^{N}_{t}$. Here, $N$ is the number of controllable agents, and other moving scene objects are treated as uncontrolled environmental dynamics. 
Previous multi-agent methods \citep{sun2026prisma,savva2026solaris,wu2026multiworld,hu2026metaworld} use primarily 2D-based world representations, suffering from degraded cross-camera and cross-agent consistency and lacking 3D geometric guidance. To tackle these problems, we reconstruct and progressively update a sparse point cloud $\mathbf{P}_{t}$ as a shared 3D state for all agents. Furthermore, our Artemis also supports multi-camera driving data through camera tiling \citep{zhu2026shareverse}, which enlarges sensor coverage for real-world applications and mitigates the potential occlusions in single-camera-only approaches.

\begin{figure}[t]
\centering
\includegraphics[width=\linewidth]{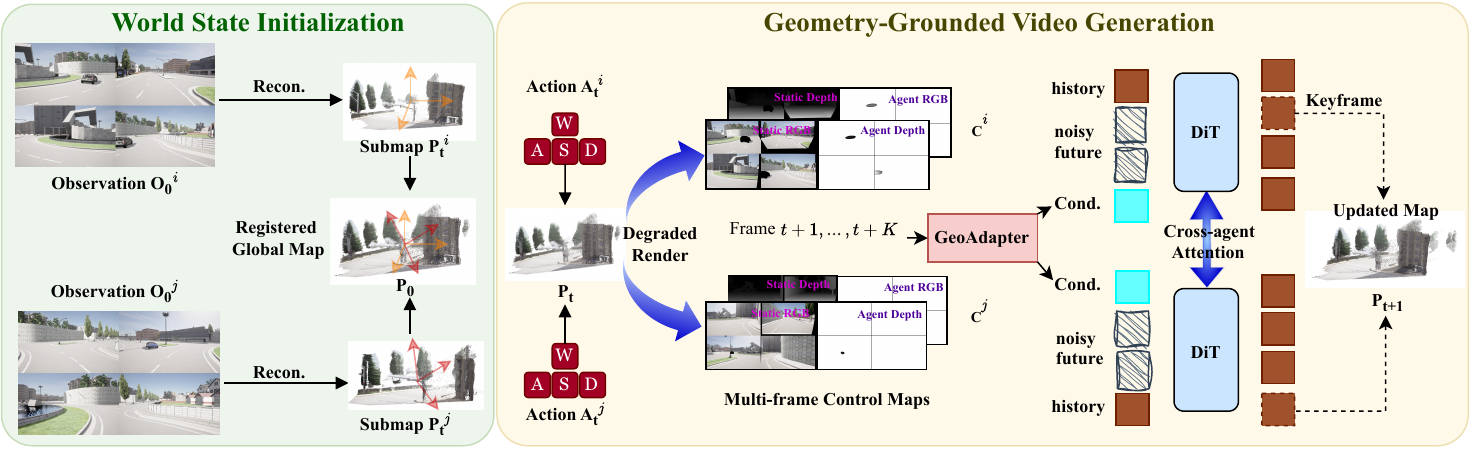}
\vspace{-6mm}
\caption{\textbf{Overall pipeline of Artemis.} We initialize the shared 3D map by registering individual observations. Given action inputs, control maps disentangling background static, foreground controlled agent proxy, and uncontrolled environmental dynamics are rendered respectively from the reconstructed global map. Then, these controls are injected into DiT, guiding the generator's rollouts. After each inference block, keyframes are selected to update the static world map. }
\vspace{-3mm}
\label{fig:pipe}
\end{figure}

\textbf{Overall Pipeline.} As in Fig. \ref{fig:pipe}, we first initialize a shared static world map from the agents' initial observations with LingBot-Map \citep{chen2026geometric}, a feed-forward foundation model for streaming 3D reconstruction, where submaps from all agents are registered in a common coordinate system. Given the input actions, we derive agent trajectories and render foreground-background disentangled control maps from the reconstructed world \citep{Yang_2026_CVPR}. Specifically, controlled-agent proxies fitted from the first frames are propagated along these trajectories, with visibility-triggered agent insertion when needed. We render the static map, controlled proxies, and available uncontrolled dynamics observation history into a unified set of controls for the planned camera views, encoding both geometry and historical dynamics. Afterwards, a GeoAdapter module injects these controls into a pre-trained video diffusion transformer, Wan2.2 5B \citep{wan2025wan}, which jointly generates synchronized future frames for all agents. Another cross attention layer \citep{savva2026solaris,wu2026multiworld} is also designed to communicate inter-agent information in latent space.  After generation, reconstructed static content updates the shared map, and accepted uncontrolled dynamics update their 3D history. Clean generated latents also update the implicit per-agent historical memories in KV-Cache for autoregressive online rollout.

\subsection{Shared 3D World State Construction}
In real world, various agents perceive and interact with the same evolving world, and the observations across agents are just different projections of the shared 3D geometry, each captured from a specific viewpoint. Motivated by this insight, we explicitly reconstruct a unified 3D world state from agents' observations and actions, ensuring that the generated videos can maintain high cross-view consistency and spatial awareness.

\subsubsection{World Initialization} 
Given the first-frame single- or multi-camera observations from all agents $\{\mathbf{O}^{n}_{0}\}_{n=1}^{N}$, we initialize the world map using a streaming feed-forward reconstruction model \citep{chen2026geometric}. Specifically, for each agent $n$, initial submaps $\mathbf{P}_{0}^{n}$ represented by static point clouds are reconstructed from their individually observed first frames and merged together to construct the shared static global map $\mathbf{P}_{0}= \cup_{n=1}^{N}(\mathbf{P}_{0}^{n})$ using the cross-agent relative pose. We exclude controlled agents, uncontrolled environmental dynamics, together with sky regions and low-confidence geometry, from static-map fusion. Furthermore, we also maintain an individual history 3D buffer $\mathcal{H}^{n}$ for each segmented uncontrolled dynamics to store their motion priors. 

\subsubsection{Action-guided Geometric Injection}
Future videos are jointly conditioned by historical observations and the input actions of each agent. To effectively control agent trajectories, we propose an action-guided geometric injection module consisting of three components: Action-conditioned Trajectory Sampling, Foreground-background Disentangled Control Signal Rendering, and GeoAdapter for latent-space geometry injection.

\textbf{Action-conditioned Trajectory Sampling.} 
During online inference, discrete action inputs can determine the future motion of the controllable agents within each rollout block. For each controllable agent $n$, an action $\mathbf{a}^n$ is represented as a local transformation $\Delta\mathbf{T}(\mathbf{a}^{n})\in\mathrm{SE}(3)$ relative to its current camera-to-world pose $\mathbf{T}_t^n$. To smoothly recover motions at any intermediate timestamp, we interpolate a sequence of intermediate trajectories from the current to the target pose $\mathbf{T}_{t}^{n}\Delta\mathbf{T}(\mathbf{a}^{n}).$  over $K$ timestamps, obtaining $\mathbf{T}_{t}^{n},\mathbf{T}_{t+1}^{n},\ldots,\mathbf{T}_{t+K}^{n}$. Translation is linearly interpolated, while rotation is interpolated on $\mathrm{SO}(3)$. 




\textbf{Foreground-background Disentangled Rendering.}
We construct the per-block control maps by independently rendering static background geometry, action-controlled foreground proxies, and history-conditioned control maps for uncontrolled dynamics. For each future pose $\mathbf{T}_{t+k}^{n}$ of the observer agent $n$, the explicit world state $\mathbf{P}_{t}$ is rendered to obtain a degraded background RGB image, a background depth map, and a visibility mask, while controllable foreground proxies are rendered to obtain foreground degraded RGB, depth, and visibility mask as:
\begin{equation}
    \mathbf{B}_{\mathrm{rgb}}^{n}, \mathbf{B}_{\mathrm{dep}}^{n}, \mathbf{M}_{\mathrm{hole},k}^{n} = \mathcal{R}_{\mathrm{bg}}\left(\mathbf{P}_t,\mathbf{T}_{t+k}^{n},\mathbf{I}^{n}\right), \quad\mathbf{F}_{\mathrm{rgb}}^{n}, \mathbf{F}_{\mathrm{dep}}^{n}, \mathbf{M}_{\mathrm{ob},k}^{n}  = \mathcal{R}_{\mathrm{ob}}\left(\mathcal{Q}_{t+k}^{\neg n},\mathbf{T}_{t+k}^{n},\mathbf{I}^{n}\right),
\end{equation}
where $\mathbf{I}^{n}$ indicates the camera intrinsics. The background-hole mask $\mathbf{M}_{\mathrm{hole},k}^{n}$ and foreground mask $\mathbf{M}_{\mathrm{ob},k}^{n}$ identifies regions that cannot be rendered by the current 3D map due to missing geometry, occlusion, or sparsity. $\mathcal{Q}_{t+k}^{\neg n}$ denotes the proxy set.

For uncontrolled dynamics, let $\mathcal{H}^{n}$ denote the historical 3D buffer available to observer $n$. At target frame $t+k$, we select observations from the buffer within a temporal window:
$
    \mathcal{S}(t+k;a)
    =
    \{s:\max(0,t+k-L+1)\leq s\leq\min(t+k,a)\},
    \label{eq:uncontrolled_window}
$
where $a=t+k$ for offline preprocessing and $a$ is the next rollout block boundary during online inference. By reprojecting these observations, we obtain:
\begin{equation}
    \mathbf{H}_{\mathrm{rgb}}^{n}, \mathbf{H}_{\mathrm{dep}}^{n},
    \mathbf{H}_{\mathrm{age}}^{n}
    =
    \mathcal{R}_{\mathrm{hist}}
    \left(\mathcal{H}_{\mathcal{S}(t+k;a)}^{n},
    \mathbf{T}_{t+k}^{n},\mathbf{I}^{n}\right),
\end{equation}
where $\mathbf{H}_{\mathrm{age}}^{n}$ is an age-weighted soft coverage map. Finally, we combine the background-hole and foreground masks together: $\mathbf{M}^{n} = \mathbf{M}_{\mathrm{hole},k}^{n}\lor\mathbf{M}_{\mathrm{obj},k}^{n}.$ The integrated mask highlights both newly exposed regions and dynamic foreground regions that still require synthesis. Overall, the complete control maps are:
$
    \mathbf{C}^{n} = \operatorname{Concat}[\mathbf{B}_{\mathrm{rgb}}^{n},\mathbf{B}_{\mathrm{dep}}^{n},\mathbf{F}_{\mathrm{rgb}}^{n},\mathbf{F}_{\mathrm{dep}}^{n},\mathbf{M}^{n}, \mathbf{H}_{\mathrm{rgb}}^{n}, \mathbf{H}_{\mathrm{dep}}^{n},
    \mathbf{H}_{\mathrm{age}}^{n}],
$
including static background RGB/depth, controllable-agent RGB/depth, the synthesis mask, and uncontrolled-history RGB/depth/age values. As for the multi-camera setting, in addition to RGB/depth controls, we also provide the DiT through token-aligned camera geometry with ray-map.



\textbf{GeoAdapter.} The rendered unified control maps are then injected into latent spaces of DiT through our GeoAdapter branch. Specifically, the RGB, depth and age controls are encoded via VAE, while the visibility mask is directly downsampled using nearest neighbor interpolation. These compressed latent controls are then embedded and injected into selected DiT layers through residual feature projections. Overall, the explicit geometry control maps can determine where proxies should appear, and the pretrained DiT correspondingly imagines the contents' appearance.

\subsection{Memory-Enhanced Streaming World Update}
To enhance cross-agent consistency and achieve progressive state update, we design three world update mechanisms for static background, uncontrolled dynamics, and agent pose, respectively.
\begin{wrapfigure}{r}{0.55\textwidth}
    \vspace{-12pt}
    \centering
    \includegraphics[width=0.54\textwidth]{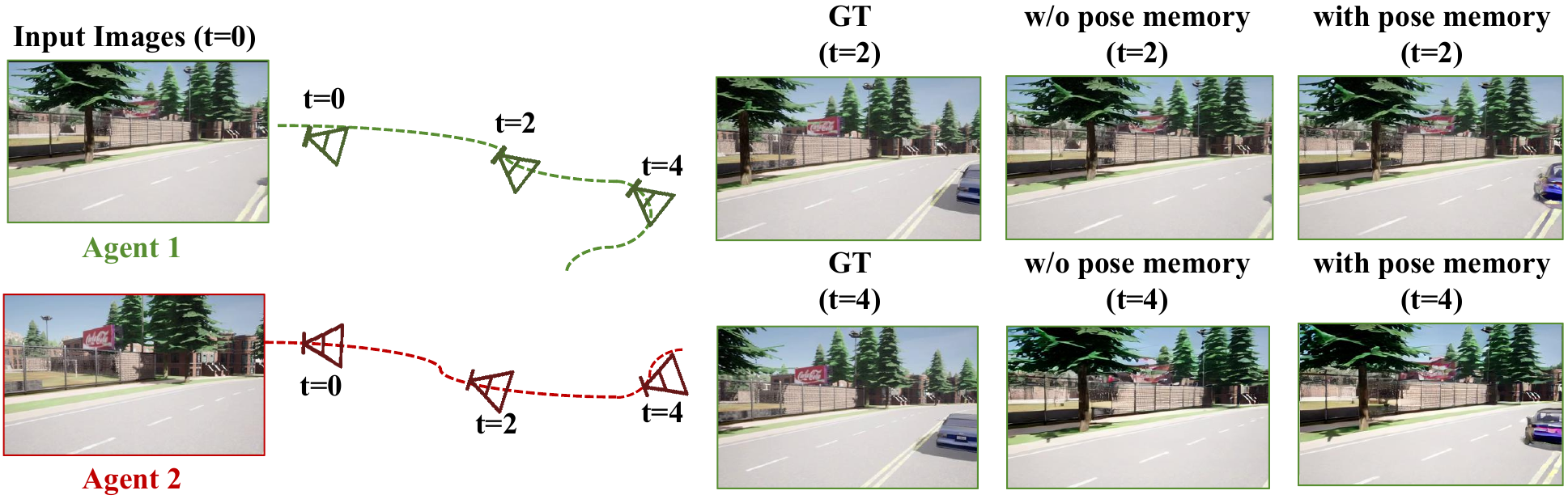}
    \vspace{-10pt}
    \caption{{Illustration of our designed pose memory.}}
    \vspace{-15pt}
    \label{fig:pos_mem}
\end{wrapfigure}
\textbf{Static Map Update.} After each rollout, we will select a few keyframes from newly generated frames in the block to update the unified 3D world map. The chosen keyframes are the frames introducing unseen scenarios or sharing low overlap with historical observations. The final world map is updated by merging the original static geometry map and the newly recovered submap reconstructed from keyframes. 

\textbf{Uncontrolled Dynamic Buffer Maintenance.} For each uncontrolled segmented dynamics, we update the history 3D buffer using a first-in-first-out principle, discarding the farthest frames while dynamically retaining the latest frame. In this case, the future positions can be extrapolated from the progressively updated historical motion priors.

\textbf{Structured Pose Memory Maintenance.}
For each controllable agent, we maintain their historical poses across rollout blocks, so that an agent can remain geometrically consistent even when it is temporarily occluded or outside the observer's sight. If one agent is not visible in the current frame, the generation model cannot generate it in the following block as original control maps fail to aware its existence through current frame segmentation. To address this issue, we design a pose-guided verification module based on maintained pose memory as in Fig. \ref{fig:pos_mem}. If no proxy is available is the current frame but the target's planned poses indicate potential entry into the observer's field of view, we initialize a conservative vehicle-shaped Gaussian and integrate it with the target position. This procedure allows the model to possess a global awareness of each agent's location and feasibly supports the generation of newly visible controlled agents at any timestamp.

\section{Experiment}

\subsection{Implementation Details}

Artemis is built on the Wan2.2-TI2V-5B DiT \citep{wan2025wan} backbone. We use a three-stage training recipe with Diffusion-Forcing \citep{chen2024diffusion} and Self-Forcing \citep{huang2025self} tailored for multi-agent online rollout. In Stage 1, we train the bidirectional model on paired CARLA videos with 15-channel geometric control maps using AdamW with an initial learning rate of $4 \times 10^{-5}$, 200 warmup steps, and cosine decay on 32 H20 GPUs. In Stage 2, we initialize from the Stage-1 checkpoint and finetune it using AdamW with a learning rate of $2 \times 10^{-5}$ and 400 warmup steps. Both stages are trained on 81-frame clips at $256 \times 416$ resolution with flow matching, 1000 training timesteps, and flow shift $\gamma = 5.0$. During online inference, we generate 81-frame videos block by block with 5-frame denoising blocks and a 33-frame history window, using 50-step Euler sampling, guidance scale $\eta = 1.0$. For stage-3 distillation, a causal generator is optimized with DMD-style distillation with 5-frame blocks and a 2-step denoising schedule.


\subsection{Quantitative Results}

\textbf{Comparison Baselines.} Since recent multi-agent world models are mostly still close-sourced, we choose a representative method Solaris \citep{savva2026solaris}, using implicit shared self-attention for cross-agent association. We also migrate a similar strategy as in OASIS \citep{decart2024oasis} by direct concatenating multi-agent tokens along feature channels.

\textbf{Evaluation Metrics.} We report eight metrics covering both appearance fidelity and multi-agent consistency. For appearance metrics, we compute PSNR, SSIM, and LPIPS between each generated video and its ground-truth video. For multi-agent consistency, we use DA3 ~\citep{lin2025depth} to process each generated video and estimate depth, camera intrinsics, and camera poses. Subject Consistency and Background Consistency evaluate the temporal stability of subject appearance and background structure, respectively. We also report the Aesthetic Quality and Imaging Quality, reflecting frame-level aesthetic appeal and overall perceptual quality, respectively.

\begin{table}[t!]
\centering
\caption{Quantitative comparisons on MA-CARLA Dataset.}
\label{tab:comparison}
\resizebox{\columnwidth}{!}{%
\renewcommand{\arraystretch}{1.08}
\begin{tabular}{lccccccc}
\toprule
Method & PSNR $\uparrow$ & SSIM $\uparrow$ & LPIPS $\downarrow$ &
Subject Consistency $\uparrow$ & Background Consistency $\uparrow$  & Aesthetic Quality $\uparrow$ &
Imaging Quality $\uparrow$ \\
\midrule
OASIS & 13.459 & 0.500 & 0.530 & 0.828 & 0.921  & 0.440 & 0.434 \\
Solaris & 10.900 & 0.378 & 0.660 & 0.759 & 0.916  & 0.392 & 0.525 \\
Solaris (finetuned) & 16.015 & 0.522 & 0.414 & 0.896 &
\textbf{0.953} & 0.502 & \textbf{0.581} \\
Ours & \textbf{17.743} & \textbf{0.566} & \textbf{0.309} &
\textbf{0.902} & 0.949  & \textbf{0.536} & 0.568 \\
\bottomrule
\end{tabular}}
\vspace{-3mm}
\end{table}

\begin{figure}[t]
\centering
\includegraphics[width=\linewidth]{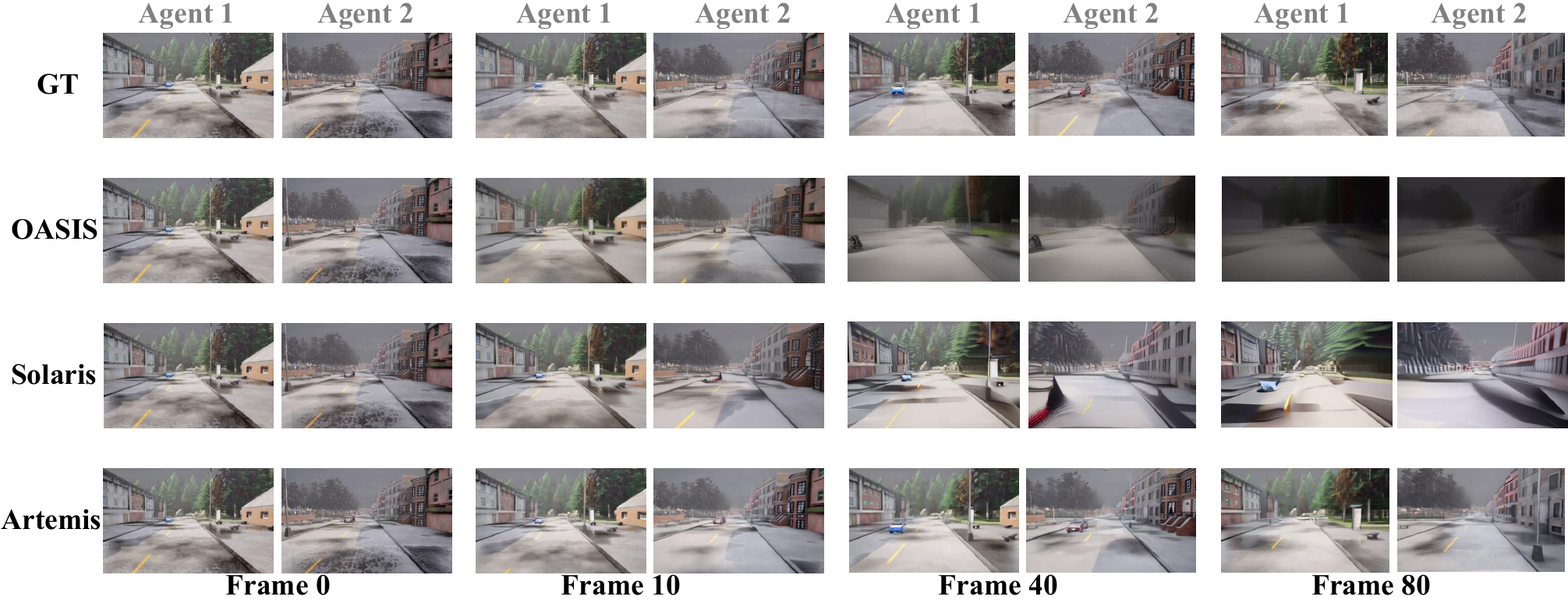}
\vspace{-6mm}
\caption{{Visualization comparisons with previous multi-agent methods.} }
\vspace{-2mm}
\label{fig:visual1}
\end{figure}


\textbf{Comparison Results.} As in Table \ref{tab:comparison}, our Artemis outperforms the baseline method Solaris and OASIS by a large margin in both appearance-related metrics and also consistency. Compared to Solaris, our Artemis achieves almost double SSIM, while reducing LPIPS to almost half. This proves that our rendered videos possess higher generation quality with an explicit geometry-guided paradigm and powerful generation priors from DiT. As for multi-agent consistency, our method also significantly surpasses Solaris in both cross-view consistency and trajectory consistency. This is mainly because implicit-only methods cannot recover underlying 3D world states, suffering from severe camera pose changes, occlusions, or few-shot views.

\subsection{Qualitative Results}
\begin{wrapfigure}{r}{0.65\textwidth}
    \centering
    \includegraphics[width=0.64\textwidth]{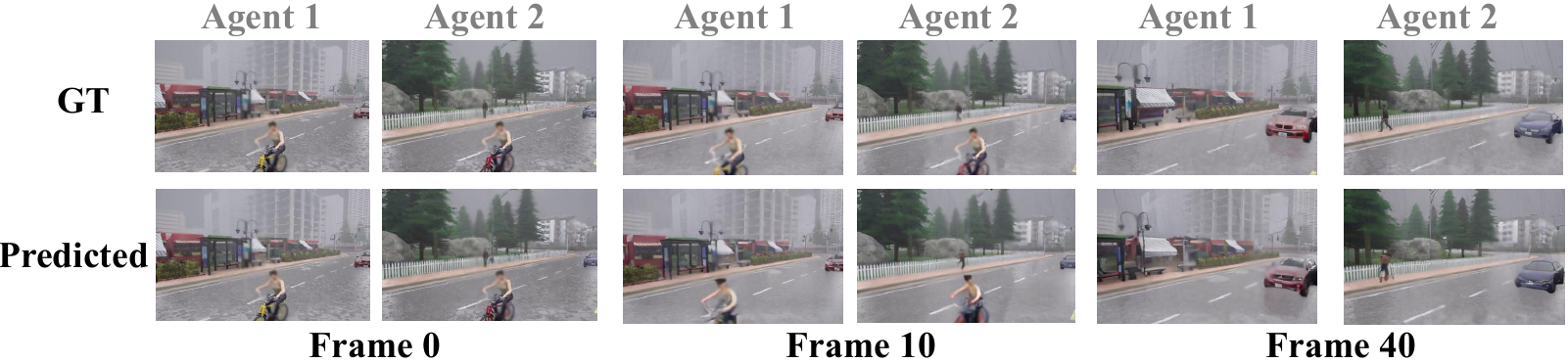}
    \vspace{-10pt}
    \caption{{Visualization of uncontrolled dynamics.}}
    \vspace{-15pt}
    \label{fig:uncontrol}
\end{wrapfigure}

\textbf{Comparison with Implicit Multi-agent World Models.} Previous multi-agent world models commonly adopt an implicit agent association manner, where cross-agent attention is introduced to exchange perspectives across agents. However, we observe that this implicit modeling can induce inconsistency geometry, especially for longer-sequence predictions. As in Fig. \ref{fig:visual1}, OASIS cannot handle the driving scenarios, with almost all blurry future imaginations. Although Solaris can generate basically reasonable geometry within the next 10 future frames, it suffers from artifact during long-horizon rollout. This illustrates that Solaris cannot learn genuine geometric consistency over time. In contrast, Artemis still forecasts accurate future videos over long sequences. We also visualize the uncontrolled dynamics in Fig. \ref{fig:uncontrol}, where two non-rigid bicyclists are predicted accurately.

\begin{figure}[t]
\centering
\includegraphics[width=\linewidth]{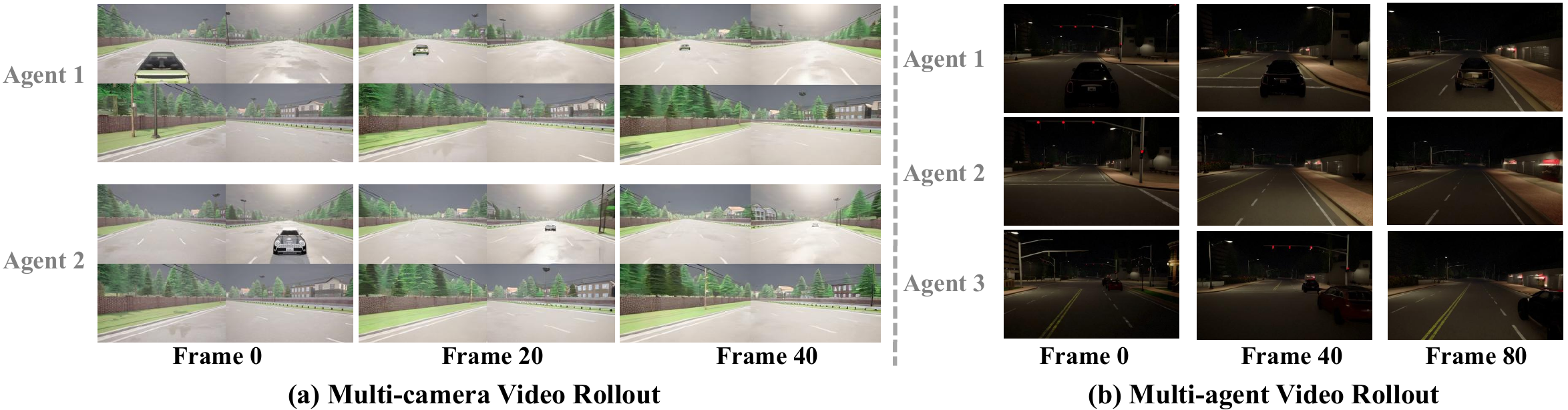}
\vspace{-6mm}
\caption{{Visualization samples in the multi-camera and multi-agent settings.} }
\vspace{-5mm}
\label{fig:visual2}
\end{figure}

\textbf{Multi-camera Rollout.} Unlike previous multi-agent world models, our collected data and training pipeline can support multi-camera settings, where more spatial ranges could be considered for better geometric reconstruction. In Fig. \ref{fig:visual2} (a), we also provide a multi-camera sample where videos samples from the directions \texttt{FRONT}, \texttt{FRONT\_LEFT}, \texttt{FRONT\_RIGHT}, and \texttt{REAR} are delivered as the conditional input. Due to the explicit reconstructed 3D state, our Artemis can still forecast cross-view consistent future videos.   



\subsection{Scalability to More Agents}

For scalability beyond two agents, we extend the pairwise pipeline by treating each vehicle as an observer and all other vehicles as controlled targets. Unlike the two-agent case, where the visible foreground object can usually be assigned to the other agent directly, the multi-agent setting requires identity-aware target matching because each observer may see multiple vehicles. For each observer, detected vehicle masks are matched to the other agents using projected trajectories, proxy-track predictions, depth consistency, and image-space distance. The matched targets are then fitted as proxy Gaussians, rolled forward with their own trajectories, and merged into one dynamic control map for that observer. Online inference results of three agents are shown in Fig.~\ref{fig:visual2} (b). 


\subsection{Ablation Studies}

\begin{table}[t]
\centering
\caption{{Ablation studies of our designed modules.}}
\label{tab:ablation1}
\resizebox{\columnwidth}{!}{%
\renewcommand{\arraystretch}{1.08}
\begin{tabular}{lccccccc}
\toprule
Method & PSNR $\uparrow$& SSIM $\uparrow$& LPIPS $\downarrow$ & Subject Consistency $\uparrow$& Background Consistency $\uparrow$ & Aesthetic Quality $\uparrow$& Imaging Quality $\uparrow$\\
\midrule
w/o proxy depth & 17.5823 &0.5613 &0.3161 &0.8982&0.9475&0.5352&\bf0.5707\\
w/o proxy RGB &17.3938& 0.5555 &0.3287&0.8966&0.9471&0.5350&0.5619\\
w/o ray-map&17.5284&0.5601&0.3263&0.8902&0.9456&0.5327&0.5635\\
w/o static map update&\bf17.7429&\bf0.5666&0.3092&0.8998&0.9486&0.5356&0.5671 \\
with separate static map&17.7236&0.5657&0.3093&0.9019&0.9494&0.5358&0.5675\\
with truncated KV-Cache&17.6488&0.5628&0.3167&0.8926&0.9465&0.5297&0.5562\\
Ours & \bf17.7429 & 0.5662 & \bf0.3085 & \bf0.9021&\bf 0.9495&\bf 0.5360& 0.5676 \\
\bottomrule
\end{tabular}}
\vspace{-6mm}
\end{table}

We conduct ablation studies to analyze the contribution of the geometric conditioning, online world update, and temporal memory mechanisms. The results are summarized in Table~\ref{tab:ablation1}.
First, removing depth-related conditions consistently degrades reconstruction quality. This confirms that the dynamic proxy depth provide complementary geometric guidance.
Second, disabling online world map update also hurts performance, reducing background and subject consistencies. This indicates that maintaining and updating the world representation during rollout is important for long-horizon consistency. Without map update, accumulated generation errors cannot be corrected by newly observed content, leading to weaker view reconstruction quality.
Reducing the KV-cache with fewer historical frames also leads to worse perceptual quality and consistency. The result suggests that the causal KV-cache is important for preserving coherent autoregressive history.

\section{Conclusion}
In this paper, we propose a novel multi-agent world model which is anchored by reconstructed 3D world state. To incorporate large-scale dynamics, we curate a new driving dataset based on the CARLA simulator. Unlike previous methods mostly leveraging implicit cross-agent correlation, our Artemis can explicitly reconstruct and dynamically update a synchronized 3D world map. We also design a history 3D buffer for uncontrolled environmental dynamics, leveraging historical observations to predict their future motion. Extensive experiments demonstrate the effectiveness of our proposed novelty, spatio-temporal consistency across frames, and scalability to more agent numbers.

\bibliography{iclr2027_conference}
\bibliographystyle{iclr2027_conference}

\appendix
\section{Appendix}
In the appendix, we supplement the additional implementation settings, data collection and processing procedures, and more experimental results. 
Overall, the structure is as follows:
\begin{itemize}
    \item More details about the dataset curation, interaction mode design, data splits, and pre-processing are provided in Section \ref{app:ma_carla}.
    \item Three-stage training procedures are introduced in Section \ref{app:training}.
   \item More visualization results are further provided in Section \ref{app:exper}.
\end{itemize}

\subsection{MA-CARLA Dataset Details}
\label{app:ma_carla}

\subsubsection{Data Collection}
\label{app:ma_carla_collection}
Each weather rendering of a trajectory counts as a scene, irrespective of camera count. Dynamic backgrounds add moving pedestrians or cyclists, which are not agents. Instrumented vehicles may move under either background setting. Before rendering, we check collisions and spacing, then assess visibility, projected size, and motion smoothness. Accepted trajectories are replayed with synchronized agents and cameras.

\subsubsection{Interaction Template Design}
\label{app:ma_carla_templates}
Templates arrange vehicles to approach from opposite directions, follow one another, or overtake in adjacent lanes. Route variants assign turns. The dynamic two-agent pool also includes braking, stop-and-go motion, and lane changes. Initial spacing and speeds vary across routes. Three-agent templates arrange opposing encounters, chases involving a parallel pair, or adjacent-lane overtakes. Each episode can contain several pairwise relationships.

\subsubsection{Figure Statistics}
\label{app:ma_carla_behaviors}
Figure~\ref{fig:dataset} summarizes recorded motion rather than generator labels. Opposing encounters are identified from relative heading and proximity. The co-directional group includes following and parallel travel, with overtaking identified by a reversal of vehicle order. The turning and crossing group covers heading changes, including road bends, or intersections of recorded paths. Spatial crossing does not require simultaneous arrival. We evaluate all vehicle pairs and count each scene once per satisfied group. Groups may overlap. Both rings combine background conditions and use 17,790 group assignments as the denominator for angles and percentages. Outer sectors divide each group by agent count. The center reports the 11,283 rendered scenes.

\subsubsection{Cameras and Data Splits}
\label{app:ma_carla_calibration}
The selected views face front, front-left, front-right, and rear, with fields of view in Table~\ref{tab:ma_carla_rigs}. Intrinsics and poses are stored per scene. Recorded timestamps determine temporal spacing, which varies across collection configurations.

\begin{table}[htbp]
\centering
\caption{Horizontal fields of view for the selected cameras.}
\label{tab:ma_carla_rigs}
\begin{tabular}{lrrr}
\toprule
Configuration & Front & Oblique (each) & Rear\\
\midrule
Two-agent static, narrow & $70^\circ$ & $70^\circ$ & $90^\circ$\\
Two-agent static, wide & $80^\circ$ & $85^\circ$ & $95^\circ$\\
Two-agent dynamic & $75^\circ$ & $75^\circ$ & $90^\circ$\\
Three-agent pool & $70^\circ$ & $70^\circ$ & $90^\circ$\\
\bottomrule
\end{tabular}
\end{table}

Table~\ref{tab:ma_carla_splits} reports the experimental splits. Multicamera experiments use the static scenes with complete camera streams. Frozen trajectory groups keep weather re-renders and all observations together within each pool. The three-agent pool is separate from the reported experiments.

\begin{table}[htbp]
\centering
\caption{Scene counts in the two-agent experimental splits.}
\label{tab:ma_carla_splits}
\begin{tabular}{lrrr}
\toprule
Setting & Train & Validation & Test\\
\midrule
Static, single camera & 4,816 & 602 & 602 \\
Static, four-view tile & 4,813 & 600 & 601 \\
Dynamic background & 2,706 & 338 & 338 \\
\bottomrule
\end{tabular}
\end{table}

\subsubsection{Scale Alignment}
In preprocessing, each agent's observation stream is reconstructed with LingBot-Map, which estimates dense depth, camera intrinsics, and camera poses in a consistent local coordinate gauge. We use these estimates jointly for background point lifting, controllable-proxy fitting, ambient-history lifting, and control-map rendering. CARLA annotations are used for identity association and motion supervision, but any metric quantities used by the renderer are converted into the LingBot-Map gauge before being combined with reconstructed geometry.

During online rollout, the initial observations define the reference scale and camera rig geometry. LingBot-Map reconstructs the initial local scene and provides the intrinsics and camera poses used for subsequent rendering and ray-map construction. User actions, inter-agent offsets, proxy extents, and generated-frame reconstructions are expressed in this same reference gauge before map fusion or dynamic-history insertion. Thus, Artemis maintains a single consistent geometric scale throughout autoregressive rollout, without requiring per-block scale re-estimation.

\subsection{Three-Stage Training Procedure}
\label{app:training}
We  design a three-stage training recipe using Diffusion-Forcing \citep{chen2024diffusion} and Self-Forcing \citep{huang2025self} and tailor them for multi-agent online rollout:

\textbf{Stage 1: Bidirectional Teacher for Geometry Grounding.} We first train a bidirectional teacher network to guarantee high-quality geometry reconstruction and effective control map injection. This stage jointly trains the cross-agent attention layer, GeoAdapter, and DiT on full sequences for all agents, using pre-processed control maps. Stage 1 can exploit larger temporal receptive field, cross-agent interactions, and spatio-temporal consistency, which serves as a real teacher for stage 3 distillation.

\textbf{Stage 2: Causal Student with Diffusion-Forcing.} Stage 2 training is initialized based on the pretrained checkpoint form stage 1. However, we switch the temporally non-causal training in stage 1 to autoregressive training with causal masks using Diffusion-Forcing \citep{chen2024diffusion}. Variant noise scales are implemented on different video frames in DiT, while the noise scales for various agents are also independently sampled. Notably, the generation process is still established on full frame sequences in a single forward pass.

\textbf{Stage 3: Self-Forcing-based Distillation.} For stage 3, we leverage online streaming rollout, autoregressively generating videos based on history self-rollouts in a chunk-wise paradigm \citep{huang2025self}, as in Sec.~\ref{inference}. This step reduces the diffusion denoising process into fewer steps which improves online inference efficiency. Specifically, control maps are first online rendered from the initialized 3D reconstructed world and intermediate actions. Then, latent frames are generated causally, where each agent will maintain an independent KV Cache to store its own historical latent frames as memory. Finally, the generated chunk is decoded into current videos, which are then used to reconstruct the updated 3D world map. We instantiate three models with identical architectures but distinct roles: a causal generator initialized from stage 2, a frozen bidirectional real teacher initialized from stage 1, and a trainable bidirectional fake critic, also initialized from stage 1. The real teacher represents the data distribution learned from the real videos, whereas the fake critic continuously tracks the evolving distribution induced by the causal generator.

\subsection{Additional Visualizations}
\label{app:exper}
\begin{figure}[t]
\centering
\includegraphics[width=\linewidth]{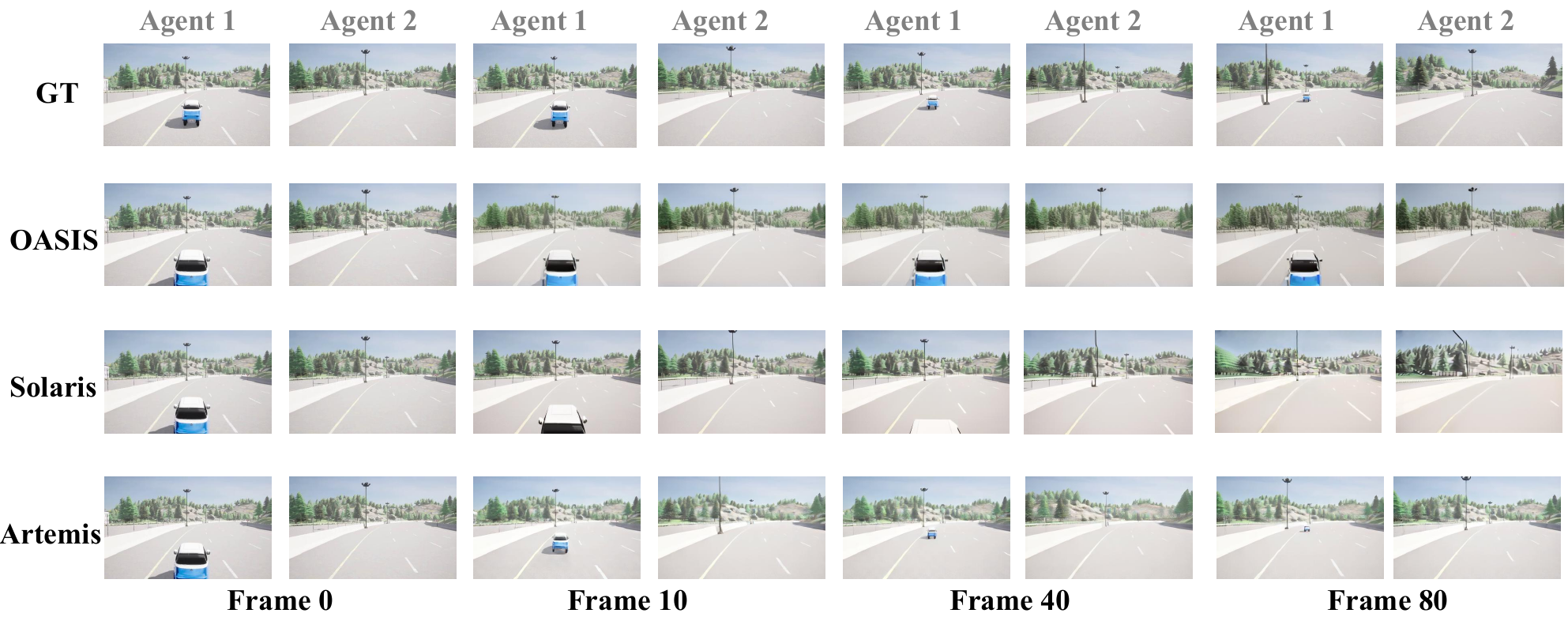}
\vspace{-6mm}
\caption{{Visualization comparisons with previous multi-agent methods.} }
\vspace{-2mm}
\label{app:visual1}
\end{figure}

We further visualize another sample in our generated videos in Fig. \ref{app:visual1}. Previous implicit methods estimate wrong agent motions, where OASIS predicts a static agent while Solaris estimates a reversed direction. Compared to them, our method can estimate accurate agent motions. Please refer to our project page \url{https://liujiuming123.github.io/Artemis/} for more visualizations.

\end{document}